\documentclass[10pt, a4paper]{article}

\usepackage{lrec-coling2024} 

\usepackage{natbib}
\usepackage{multibib}
\makeatletter
\def\@mb@citenamelist{cite,citep,citet,citealp,citealt,citepalias,citetalias}
\makeatother
\newcites{languageresource}{~}

\usepackage{graphicx}
\usepackage{tabularx}
\usepackage{multirow}
\usepackage{soul}

\usepackage{url}
\usepackage{hyperref}
\usepackage[hyphenbreaks]{breakurl}

\hypersetup{breaklinks=true}

\usepackage{xstring}

\usepackage{color}

\title{Assessing the Efficacy of Grammar Error Correction: A Human Evaluation Approach in the Japanese Context}

\name{Qiao Wang\textsuperscript{1} and Zheng Yuan\textsuperscript{2}} 

\address{
\textsuperscript{1} Center for English Language Education, Faculty of Science and Engineering (CELESE),\\ Waseda University, Tokyo, Japan \\
\textsuperscript{2} Department of Informatics, King’s College London,
         U.K.  \\
        judy.wang@aoni.waseda.jp, zheng.yuan@kcl.ac.uk, \\
       }

\abstract{
In this study, we evaluated the performance of the state-of-the-art sequence tagging grammar error detection and correction model (SeqTagger) using Japanese university students' writing samples. With an automatic annotation toolkit, ERRANT, we first evaluated SeqTagger's performance on error correction with human expert correction as the benchmark. Then a human-annotated approach was adopted to evaluate Seqtagger's performance in error detection using a subset of the writing dataset. Results indicated a precision of 63.66\% and a recall of 20.19\% for error correction in the full dataset. For the subset, after manual exclusion of irrelevant errors such as semantic and mechanical ones, the model shows an adjusted precision of 97.98\% and an adjusted recall of 42.98\% for error detection, indicating the model’s high accuracy but also its conservativeness. Thematic analysis on errors undetected by the model revealed that determiners and articles, especially the latter, were predominant. Specifically, in terms of context-independent errors, the model occasionally overlooked basic ones and faced challenges with overly erroneous or complex structures. Meanwhile, context-dependent errors, notably those related to tense and noun number, as well as those possibly influenced by the students' first language (L1), remained particularly challenging.
 \\ \newline \Keywords{Grammar error correction, Grammar error detection, Japanese university students, Human evaluation, Language learning} }

\begin{document}

\maketitleabstract

\section{Introduction}

Grammatical error correction (GEC) is the
task of automatically detecting and correcting grammatical errors in text, which has important pedagogical applications that can benefit both students and teachers in classroom, as well as online language learning. GEC has often been considered a sequence-to-sequence translation task, where systems learn to ``translate'' an ungrammatical input sentence to a grammatical output one \citep{yuan-briscoe-2016-grammatical,kiyono-etal-2019-empirical,yuan-etal-2019-neural,TACL2047}. More recently however, \citet{omelianchuk-etal-2020-gector} proposed a sequence tagging approach for GEC that can perform both error detection and correction, and benefit from fast inference compared to sequence-to-sequence GEC models. Following this trend, several sequence tagging GEC systems have been developed \citep{mesham-etal-2023-extended}, and state-of-the-art results on public benchmark datasets have been reported using automatic GEC evaluation metrics like ERRor ANnotation Toolkit (ERRANT) \citep{bryant-etal-2017-automatic}, M\textsuperscript{2} Scorer~\citep{dahlmeier-ng-2012-better} and GLEU \citep{napoles-etal-2015-ground}. However, much less work has been done on human evaluation of the efficacy of such GEC models in the context of language education with a linguistic focus.

In this study, we employed the state-of-the-art SeqTagger\footnote{\url{https://github.com/StuartMesham/gector_experiment_public}} GEC model from \citet{mesham-etal-2023-extended}, which introduced new transformation tags to help simplify the sequence tagging problem, as well as improve the generalisation of the GEC system. A fully annotated learner dataset with Japanese as the first language (L1) has been created to help evaluate the model's efficacy. A new human evaluation scheme has been proposed where a human-annotated approach was adopted to highlight agreement and disagreement between the errors detected by both human experts and GEC models. Further thematic analysis was employed to identify errors that are still challenging for state-of-the-art GEC models. The new dataset and human evaluation data is made publicly available to facilitate future research.\footnote{\url{https://doi.org/10.6084/m9.figshare.25295743.v1}}

\begin{table*}[t!]
\centering
\begin{tabular}{|l|c|c|c|c|c|c|c|}
\hline
& \textbf{No. of sentence pairs} & \textbf{TP} & \textbf{FP} & \textbf{FN} & \textbf{P} & \textbf{R} & \textbf{F\textsubscript{0.5}} \\ \hline
Full dataset & 1,577 & 820         & 468         & 3,242        & 63.66        & 20.19       & 44.50         \\ \hline
Human evaluation dataset & 300 & 242         & 5         & 321        & 97.98        & 42.98       & 78.01         \\ \hline

\end{tabular}
\caption{\label{performance}Performance metrics of SeqTagger with human corrections as the reference}
\end{table*}


\section{Dataset}
The Japanese university students’ writing sample dataset was sourced from 71 sophomore students majoring in science and engineering at a Japanese university. Their English proficiency levels ranged from higher B1 to higher B2 according to the CEFR framework.\footnote{\scriptsize{\url{https://www.coe.int/en/web/common-european-framework-reference-languages}}} These participants were distributed across two distinct classes and were enrolled in a year-long Academic Reading course\footnote{\url{https://celese.jp/courses/undergraduate/ar1/}
} spanning two semesters. As a course requirement, students were tasked with composing two reaction papers each semester. In these papers, students provided summaries and critical reflections on science news articles of their choosing. In total, 261 writing samples were gathered.


\section{Grammar Error Analysis}
To assess the efficacy of the state-of-the-art SeqTagger GEC model, we first engaged human experts alongside SeqTagger to detect and correct grammatical errors in the provided writing samples. 
Subsequently, an automated comparative analysis was conducted between the corrections made by human experts and those generated by the model.
In addition, it is crucial to recognize that multiple correction methods can address a single grammar error. Therefore, we also looked specifically at how the model agreed with human experts in error detection through human annotation. 

\subsection{Human and model correction}

Two native-speaking teachers meticulously examined each writing sample, selecting sentences with grammar errors (``original sentences''). They were provided with examples differentiating between grammar and semantic errors, and were instructed to avoid choosing sentences based on semantic ambiguity. Then they were asked to correct grammatical mistakes in the chosen sentences with minimal syntactic alterations. Examples of revision were also given. Each teacher was assigned half the 261 samples. 

Recognizing the inherent subjectivity in grammatical judgments \citep{10.1162/coli_a_00478}, we acknowledged that even experts might differ on certain grammar interpretations. To ensure alignment, the teachers submitted the first 100 sentences for feedback from the research team before continuing. After they finished selecting the sentences and correcting the errors, they cross-checked each other's work, noting disagreements or overlooked errors. They then integrated each other's feedback and submitted their revised selections and corrections to the researchers. A third expert was then engaged to review and finalize the sentences and corrections (``corrected sentences''). Ultimately, 1577 sentence pairs, consisting of original and corrected sentences, were produced.

The automated correction phase involved inputting the same original sentences into the SeqTagger model. This produced an equivalent number of machine-corrected sentence pairs. Consequently, two distinct datasets emerged: the human dataset and the machine dataset.

\subsection{Comparative analysis scheme}
For both datasets, we used ERRANT to align the original and corrected sentences and tag the errors. Output files in M2 format were created, locating every correction made in each sentence pair. The following is an example from the machine dataset:
\vspace{4pt}

\textit{Original:} It is surprising to know that the latest 
games have \textcolor{red}{high} technologies which enable to 
connect games and physical activities.

\textit{Corrected:} It is surprising to know that the latest 
games have \textcolor{green}{advanced} technologies which enable \textcolor{green}{us} 
to connect games and physical activities.  

\vspace{8pt}

\textit{Output M2:\footnote{Here, lines starting with `S' represent original sentences, while those starting with `A' indicate ERRANT-identified edits. Each `A' line delineates the beginning and end token offsets of the modification, the type of error, and the suggested correction in tokenized form. The subsequent fields are historically retained. When no correction is made, the output is a ``noop'' edit signifying the system found no necessary changes to the original sentence \citep{dahlmeier-ng-2012-better}.}}
\begin{small}
\begin{verbatim}
S It is surprising to know that the latest 
games have high technologies which enable 
to connect games and physical activities.

A 10 11|||R:ADJ|||advanced|||REQUIRED|||-NONE-|||0
A 14 14|||M:PRON|||us|||REQUIRED|||-NONE-|||0    
\end{verbatim}
\vspace{4pt}

\end{small}

To evaluate the model's performance in error correction, with the human dataset as the benchmark, we 
quantified the model's true positives (TF), false positives (FP), false negatives (FN), precision (P), recall (R), and F\textsubscript{0.5} scores under each error category based on the M2 files.

The evaluation of the models' performance in error detection required human annotation of agreement and disagreement between human and model detection. Anticipating an intensive workload for the in-depth qualitative analysis, a decision was made to limit the evaluation to a subset of the complete dataset, which included 300 sentences. Two annotators put the ERRANT data for both subsets side by side to ascertain whether an error detected by humans was identified or overlooked by the model, and vice versa. To offer a comprehensive analysis, categories and subcategories were introduced. Disagreements between human and model detection were further annotated with comments, which were used in subsequent thematic analysis to provide insights into the limitations of the model, as well as those of human detection.

\section{Results}

ERRANT automatic evaluation results for the performance metrics of the model (SeqTagger) with human corrections as the reference on the full dataset, including true positives (TPs), false positives (FPs), false negatives (FNs), precision (P), recall (R) and the F\textsubscript{0.5} score are presented in Table \ref{performance}. The state-of-the-art SeqTagger GEC model achieved an F\textsubscript{0.5} score of 44.50\% on our Japanese university students’ writing sample dataset.\footnote{We notice that the results here are worse than those reported on public benchmarks (e.g. BEA-dev and BEA-test~\citep{bryant-etal-2019-bea}) by \citet{mesham-etal-2023-extended}, as the model has not been trained or tuned for Japanese L1.}

For the selected 300 sentence pairs for human annotation, based on our analysis scheme, errors were deductively categorized into two main groups: ``Agreed Detections'' and ``Disagreed Detections''. Additionally, we introduced a category called ``Excluded Items'' to account for cases that didn't fit within the primary two classifications. Agreed Detections encompasses errors identified by both human annotators and the model, while Disagreed Detections represents errors detected exclusively by either entity. We further created categories and subcategories for a more nuanced understanding of the natures of errors. Three subcategories were also generated under Excluded Items: no corrections, repeated corrections and collateral corrections, and they have been excluded from the analysis to avoid overestimation.\footnote{No corrections: 18 instances; Repeated corrections: 30 instances; Collateral corrections: 108 instances} In the following, we 
present the results for agreed and disagreed detections in detail.

\subsection{Human-model agreement}
 Four major categories under agreed detections between human experts and the model were generated, including context-independent grammar errors, context-dependent grammar errors, semantic errors and mechanical errors. Table \ref{agreement} shows the results for agreed detections and the detailed explanations of the categories and subcategories are laid out below.

\begin{table}[t]
\centering
\resizebox{\columnwidth}{!}{\begin{tabular}{|l|l|c|}
\hline
\textbf{Categories} & \textbf{Subcategories} & \textbf{Instances} \\
\hline
Context-independent grammar errors & Same revisions & 202 \\
\cline{2-3} 
 & Different revisions & 39 \\
\hline
Context-dependent grammar errors & & 1 \\
\hline

\multicolumn{2}{|l|}{\textbf{Subtotal of grammar errors}} & 242 \\
\hline

\hline
Semantic errors & & 1 \\
\hline
Mechanical errors & & 2 \\
\hline

\hline
\multicolumn{2}{|l|}{\textbf{Total}} & 245 \\
\hline
\end{tabular}}
\caption{\label{agreement}Agreed Detections}
\end{table}

\paragraph{Context-independent grammar errors} are those that are inherently incorrect, irrespective of their surrounding context. 
This category is further divided into ``same revisions'' and ``different revisions''. The former refers to situations where both human experts and the model converged on identical corrections. 
Meanwhile, the latter refers to cases where the model and human experts adopted distinct corrections that are both acceptable for an agreed detection. 





\paragraph{Context-dependent grammar errors} hinge on context for accurate judgment of grammaticality. Without the specific context, they might pass as grammatically accurate. 

\paragraph{Semantic errors} target the essence or meaning of words without referencing grammatical accuracy, e.g. both humans and the model changed ``scene'' to ``scenery''. 

\paragraph{Mechanical errors} were punctuation and spelling errors. 

\subsection{Human-model disagreement}

\begin{table*}[t]
\resizebox{\textwidth}{!}{\begin{tabular}{|ll|c|c|}
\hline
\multicolumn{1}{|l|}{\textbf{Categories}}                      & \textbf{Subcategories} & \multicolumn{1}{l|}{\textbf{Undetected Instances (S)}} & \multicolumn{1}{l|}{\textbf{Undetected Instances (M)}} \\ \hline
\multicolumn{1}{|l|}{Context-independent grammar errors}     & & 196 & 3 \\ \hline
\multicolumn{1}{|l|}{Context-dependent   grammar errors}       &                        & 125                                                    & 1                                                      \\ \hline

\multicolumn{2}{|l|}{\textbf{Subtotal of grammar errors}}                               & 321                                                    & 4      
\\ \hline

\hline

\multicolumn{1}{|l|}{\multirow{2}{*}{Unnecessary corrections}} & Original correct       & 23                                                     & 4                                                      \\ \cline{2-4} 
\multicolumn{1}{|l|}{}                                         & Debatable corrections  & 29                                                     & 1                                                      \\ \hline 
\multicolumn{2}{|l|}{\textbf{Subtotal of unnecessary corrections}}                          & 52                                                     & 5                                                      \\ \hline

\hline
\multicolumn{1}{|l|}{Semantic errors}                          &                        & 113                                                    & 11                                                     \\ \hline
\multicolumn{1}{|l|}{Mechanical errors}                        &                        & 45                                                     & 6                                                      \\ \hline  

\hline
\multicolumn{2}{|l|}{\textbf{Total}}                                                    & 531                                                    & 26                                                     \\ \hline
\end{tabular}}
\caption{\label{disagreement} Disagreed Detections}
\end{table*}

Annotation results of human-model disagreement resembled the categories formed under agreement, namely: context-independent grammar errors, context-dependent grammar errors, semantic errors, and mechanical errors. However, upon closer inspection, we also observed instances where disagreements were caused by unnecessary corrections made either by humans or the model. Table \ref{disagreement} shows the annotation results. In the table, the column labeled ``Undetected Instances (S)'' quantifies the occasions where the  model failed to identify an error detected by human experts. Conversely, the column ``Undetected Instances (M)'' counts the scenarios where human experts overlooked errors that the model successfully detected.


\paragraph{Unnecessary corrections} were further sub-divided into ``original correct'' and ``debatable corrections''. The ``original correct'' subcategory pertains to situations where an original sentence was accurate upon closer inspection, but was corrected extraneously by either humans or the model. 
The ``debatable corrections'' subcategory captures errors deemed borderline: while they might violate strict grammatical conventions, they align with evolving language norms. 

\subsection{Adjusted performance metrics on the human annotation subset}
For the complete dataset, the performance metrics were calculated based on error correction. In other words, a true positive is an error identified and corrected exactly in the same way as the human experts have done. For the human annotation subset, as mentioned earlier, the focus is on detection. Thus, upon obtaining the quantitative data reflecting agreements and disagreements in terms of detection, 
 we adjusted the model's metrics, accounting for both context-dependent and context-independent grammatical errors. Results show that the model achieved an F\textsubscript{0.5} score of 78.01\%, with 97.98\% precision and 42.98\% recall for error detection, strikingly different from the metrics based on error correction (see Table \ref{performance}).  

\section{Discussions}
\subsection{Quantitative results}
Based on results regarding semantic errors and unnecessary correction, it can be seen that human experts often incorporate semantic adjustments to complement grammatical corrections, evidenced by the significant count of semantic corrections (n=113 out of 531). Contrarily, the model infrequently detect these semantic nuances. In addition, human experts occasionally ``over-correct'' or introduce unnecessary grammatical adjustments (n=52 out of 531).

Perhaps most importantly, the model exhibits a high precision in error detection, as almost every error detected by the model was a true error. However, the recall is less than 50\%, caused by the model's notable deficiency in detecting both context-dependent and context-independent grammar errors. In particular, context-dependent errors remain a challenging domain for model detection, with only 1 successfully detected by the model and 125 went unnoticed.

The impact of precision and recall varies between students and teachers. For students, a high precision is preferred to avoid misleading them with false detection. This indicates that the current model is a readily applicable tool for Japanese students. However, for teachers, a higher recall is preferable, as it reduces the time needed for error identification, allowing them to focus more on assessing the correctness of detected errors. Therefore, our next step will be to improve the model's recall to better support teachers in providing thorough feedback. 

\subsection{Thematic analysis on grammar errors undetected by the model}
For grammar errors undetected by the model, including both context-dependent and context-independent ones, we added comments to each of such cases. The following themes emerged from the comments:

\paragraph{Determiners and Articles} The model consistently underperforms in the domain of determiners and, more specifically, articles. A potential reason is the significant challenge that English articles present to Japanese students \cite{izumi-etal-2003-automatic} due to the lack of equivalent constructs in their native language \cite{snape2016teaching}. This poses great challenges to the model as it was not trained specifically for the population. Another reason may be the inherent intricacy of article usage, with studies underscoring that even native speakers occasionally display disagreements in this domain \cite{chodorow2010utility}. The difficulty in automatically detecting article errors was also highlighted in previous studies \cite{nagata-etal-2005-detecting}.

\paragraph{Noun Number} Errors in noun numbers often intertwine with those of articles and determiners, as converting a singular noun to plural can sometimes rectify the absence of an article or determiner preceding a singular noun \cite{berend2013lfg}. However, a deeper issue might stem from the Japanese language's lack of obligatory plural marking for nouns \cite{snape2016teaching}, and the resulting frequent errors among Japanese students were beyond the detection capability of the model. Further complicating matters are nouns that can be both countable and uncountable depending on their sense or metonymic usage, such as ``coffee'' which denotes an uncountable substance or a countable cup of the beverage 
 \cite{berend2013lfg}. Discriminating between these contexts is a challenging task, demanding human judgment or sophisticated NLP techniques.

\paragraph{Tense} Tense errors were heavily context-dependent and conspicuous in their absence from the model's detections. The context-dependent nature of such errors was supported by \citet{granger1999uses}, who conducted a comprehensive analysis of two error-tagged learner corpora on tense use and found that there would be no tense error if the segment was taken out of context. Further, Granger's study revealed that the simple present and simple past tenses accounted for 32.5\% and 29\% of all verb tense errors, also echoing the findings of this study. 

\paragraph{Miscellaneous} Though not a major theme, subject-verb agreement also caught our attention, as some cases are straightforward and should have been easily detected by the model. For instance, in the sentence ``\textit{If it occur human, it is very terrible.}'', the verb ``occur'' should conform to third person singular agreement following the pronoun ``it''. However, the model missed this basic error. 

Additionally, there are cases where the intricacy or elongated nature of a sentence seems to impede detection. Consider the example, ``\textit{The orbiting line that is caused the objects are sucked seriously look like noodles, so scientists say it spaghettification.}'' Here, ``look'' should indeed be ``looks'', but given the sentence's complexity and presence of other errors, the model overlooked the subject-verb agreement issue. 

\section{Conclusion}

In this study, we compared human and model detection of grammar errors in Japanese university students' writing samples and conducted qualitative analysis on the errors undetected by the state-of-the-art SeqTagger GEC model. Quantitative results show a high adjusted precision (97.98\%) but a relatively low adjusted recall (42.98\%) of the model in error detection, indicating an accurate but conservative approach. A significant revelation from our thematic analysis is the model's deficiency in detecting issues with determiners and particularly articles in both context-dependent and context-independent errors. In terms of context-independent errors, basic ones such as subject-verb agreement remain areas where the model can benefit from further refinement. There were also instances where original sentences were laden with errors or possessing elongated structures, leading the model to struggle in identifying these errors. More importantly, context-dependent errors pose a substantial challenge, such as tense and noun number. GEC models like SeqTagger, lacking the innate human cognitive ability to infer context, struggle to match human performance in this area. 





\section*{Ethical Considerations}

Subjectivity in determing grammar errors is a major ethical consideration in this study. While measures were taken to ensure a consistent grammar error detection process, subsequent analysis by the researchers highlighted instances of corrections that were deemed unnecessary. This suggests that personal biases in comprehending and interpreting grammar are not just possible but inevitable, even among experts. 

For data privacy, every precaution was taken to ensure the data's anonymity, making sure no personally identifiable information was retained. Moreover, it is worth noting that students had provided consent for their writings to be utilized, albeit only in fragmented sentences and not in their entirety. 


\section*{Limitations}
A salient limitation in this research is the inevitable subjectivity associated with grammar judgments. By its very nature, grammar is interpretative, leading even seasoned experts to possibly diverge in their evaluations or opinions about certain grammatical aspects. The research did involve a thorough cross-examination between the two experts, and the inclusion of a third expert to provide insights. Yet, this doesn't completely overcome the subjective nature intrinsic to grammatical evaluations. This is also the reason why we did not report inter-rater agreement between the two experts. 

Furthermore, no thematic analysis was conducted on the instances of agreement between the model and human experts. We did not compare the detected and undetected cases that may fall under the same theme to examine why the model sometimes identifies and at other times fails to detect the same type of error. This oversight might conceal certain patterns or insights regarding the model's performance.

The sample size is another limitation of the study. With the full dataset comprising only 261 writing samples and a smaller subset of 300 sentences selected for human annotation, the scope of our findings may have been restricted. This limitation stems from the extensive time required for thorough human evaluation. Moving forward, we aim to expand our research by incorporating a larger collection of writing samples. To facilitate this expansion and alleviate the burden on human evaluators, we plan to employ large language models as an auxiliary tool in the validation process.

 Another aspect to consider is the generalizability of the study. This study used writing sample from a single population, the Japanese student cohort. Thus, it might not fully encapsulate linguistic nuances inherent to various cultural or linguistic backgrounds.





%
\nocite{*}
\section*{Bibliographical References}\label{reference}

\bibliographystyle{lrec-coling2024-natbib}
\bibliography{lrec-coling2024-example}




\end{document}